\documentclass[conference,a4paper]{IEEEtran}
\IEEEoverridecommandlockouts

\usepackage{fancyhdr}
\usepackage[dvips]{graphicx}
\usepackage{url}
\usepackage{amsmath}
\usepackage{subfigure}
\usepackage{multirow}

\ifCLASSINFOpdf
\else
\fi

\begin{document}
\title{Re-learning of Child Model for Misclassified data by using KL Divergence in AffectNet: A Database for Facial Expression
\thanks{\copyright 2019 IEEE. Personal use of this material is permitted. Permission from IEEE must be obtained for all other uses, in any current or future media, including reprinting/republishing this material for advertising or promotional purposes, creating new collective works, for resale or redistribution to servers or lists, or reuse of any copyrighted component of this work in other works.}
}
\author{\IEEEauthorblockN{Takumi Ichimura}
\IEEEauthorblockA{Advanced Artificial Intelligence Project Research Center,\\
Research Organization of Regional Oriented Studies,\\
and Faculty of Management and Information System,\\
Prefectural University of Hiroshima\\
1-1-71, Ujina-Higashi, Minami-ku, \\
Hiroshima 734-8558, Japan\\
E-mail: ichimura@pu-hiroshima.ac.jp}
\and
\IEEEauthorblockN{Shin Kamada}
\IEEEauthorblockA{Advanced Artificial Intelligence Project Research Center,\\
Research Organization of Regional Oriented Studies,\\
Prefectural University of Hiroshima\\
1-1-71, Ujina-Higashi, Minami-ku, \\
Hiroshima 734-8558, Japan\\
E-mail: skamada@pu-hiroshima.ac.jp}
}

\maketitle
%\thispagestyle{plain}

%\fancypagestyle{plain}{
%\fancyhf{}	% clear all header and footer fields
%\fancyhead[L]{{\bf 2019 IEEE 11th International Workshop on Computational Intelligence and Applications\\ November 9-10, 2019, Hiroshima, Japan}}
%
%
%\fancyfoot[L]{{\bf 978-1-7281-2429-2/19/\$31.00 ~\copyright 2019~IEEE}}
%
%\fancyfoot[C]{}
%\fancyfoot[R]{}
%\renewcommand{\headrulewidth}{0pt}
%\renewcommand{\footrulewidth}{0pt}
%}

\pagestyle{fancy}{
\fancyhf{}
\fancyfoot[R]{}}
\renewcommand{\headrulewidth}{0pt}
\renewcommand{\footrulewidth}{0pt}

\begin{abstract}
  AffectNet contains more than 1,000,000 facial images which manually annotated for the presence of eight discrete facial expressions and the intensity of valence and arousal. Adaptive structural learning method of DBN (Adaptive DBN) is positioned as a top Deep learning model of classification capability for some large image benchmark databases. The Convolutional Neural Network and Adaptive DBN were trained for AffectNet and classification capability was compared. Adaptive DBN showed higher classification ratio. However, the model was not able to classify some test cases correctly because human emotions contain many ambiguous features or patterns leading wrong answer which includes the possibility of being a factor of adversarial examples, due to two or more annotators answer different subjective judgment for an image. In order to distinguish such cases, this paper investigated a re-learning model of Adaptive DBN with two or more child models, where the original trained model can be seen as a parent model and then new child models are generated for some misclassified cases. In addition, an appropriate child model was generated according to difference between two models by using KL divergence. The generated child models showed better performance to classify two emotion categories: `Disgust' and `Anger'.
\end{abstract}


% \fi
\IEEEpeerreviewmaketitle

\section{Introduction}
Deep learning techniques have been studied as the most advanced AI researches to various kinds of problems. An optimal architecture of network is required to reach high capability, but the development at new domain is a difficult work even for a skilled researcher. The adaptive structural learning method of Deep Belief Network (Adaptive DBN) \cite{Kamada18_Springer} has an outstanding function of determination for the network structure of Restricted Boltzmann Machine (RBM) \cite{Hinton06,Hinton12} which is the self-organized process by hidden neuron generation and deletion algorithm during learning phase. The algorithm monitors the training situation of some parameters at the RBM network. Adaptive DBN is the hierarchical model of RBMs where a new RBM is also automatically generated to monitor the total error of deep learning. Adaptive DBN method shows the highest classification capability for image recognition of some benchmark data sets such as MNIST \cite{LeCun98a}, CIFAR-10, and CIFAR-100 \cite{CIFAR10}. The classification accuracy for training data sets has been reached almost 100\% and 99.5\%, 97.4\%, and 81.2\% for test cases, respectively. The classification ratio reported in \cite{Kamada18_Springer} is higher than that of CNN (Convolutional Neural Network) such as AlexNet \cite{AlexNet}, GoogLeNet \cite{GoogLeNet}, VGG16 \cite{VGG16}, and ResNet \cite{ResNet}.

Recently, some databases of facial expression and emotion attracted much attentions. Mera et al. has proposed the Emotion Generating Calculations (EGC) which is a method to calculate an agent's emotion from the contents of utterances and to express emotions which are aroused in computer agent by using synthesized facial expression \cite{Ichimura03, Mera05}. EGC \cite{Mera03} based on the Emotion Eliciting Condition Theory \cite{Elliott92} can decide whether an event arouses pleasure or not and quantify the degree under an event. However, many facial image data were not collected to build the network with sufficient classification capability and then the trained neural network had only poor representation power.

Such models to quantify affective facial behaviors are classified into three categories: ``categorical model,'' ``dimensional model,'' and ``Facial Action Coding System (FACS) model.'' These databases are captured from movies or websites. In the categorical model, emotion is chosen from affective categories such as six basic emotions by Ekman et al. \cite{Ekman69}. The categorical model cannot explain mixture of emotions by using the simple combination of emotional words such as ``happily surprised''. As the second model, the dimensional model has valence and arousal. The facial images plotted in the 2D space of valence and arousal. The valence refers to how positive or negative an event is, and the arousal reflects whether an event is exciting or calm \cite{Russell85}. Facial Action Coding System (FACS) model has all possible facial actions in Action Units (AUs) \cite{Ekman78}. However, FACS model does not explain the affective state directly, although it shows the facial movements.

AffectNet \cite{AffectNet} is a kind of the dimensional model which can deal both intensity of emotion and different emotion categories in the continuous dimensional model. AffectNet contains more than one million images with faces and extracted facial landmark points. In \cite{AffectNet}, 12 human experts manually annotated 450,000 of these images in both categorical and dimensional (valence and arousal) models.

The developed deep learning model in \cite{AffectNet} does not become to perform good classification capability (72.0\%). On the contrary, Adaptive DBN shows almost 100.0\% for the training data set, but it remains to be around 80.0\% for the specified category of the test data because human emotions contain many ambiguous features or patterns leading wrong answer which includes the possibility of being a factor of adversarial examples \cite{Nguyen15}, due to two or more annotators answer different subjective judgment for an image. In order to represent such cases, two or more child models are investigated in addition to the parent model. We developed a new child model to train only for samples that classification accuracy is not much high and investigated the Kullback-Leibler (KL) Divergence between the original parent model and the child model by Adaptive DBN. As an experimental result, we had the situation that requires an additional child model according to KL divergence. This paper discusses the relation between two or more deep learning models by using KL divergence.

The remainder of this paper is organized as follows. In section \ref{sec:Adaptive_dbn}, basic idea of the adaptive structural learning of DBN is briefly explained.  The section \ref{sec:AffectNet} explains the facial database: AffectNet. In section \ref{sec:Experiments}, the effectiveness of our proposed method is verified on AffectNet. In section \ref{sec:Conclusion}, we give some discussions to conclude this paper.

\section{Adaptive Structural Learning Method of Deep Belief Network}
\label{sec:Adaptive_dbn}

The basic idea of our proposed Adaptive DBN is described to deepen understanding the sophisticated method.

\subsection{Neuron Generation and Annihilation Algorithm of RBM}
\label{subsec:adaptive_rbm}
While recent deep learning model has higher classification capability, the size of its network structure or the number of its parameters that a network designer must determine are larger. For the problem, we have developed the adaptive structural learning method in RBM model, called Adaptive RBM \cite{Kamada18_Springer}. RBM is an unsupervised graphical and energy based model on two kinds of layers; visible layer for input and hidden layer for feature vector, respectively. The neuron generation algorithm of Adaptive RBM is able to generate an optimal size of hidden neurons for given input space during its training situation.

\begin{figure}[]
\begin{center}
\subfigure[Neuron generation]{\includegraphics[scale=0.5]{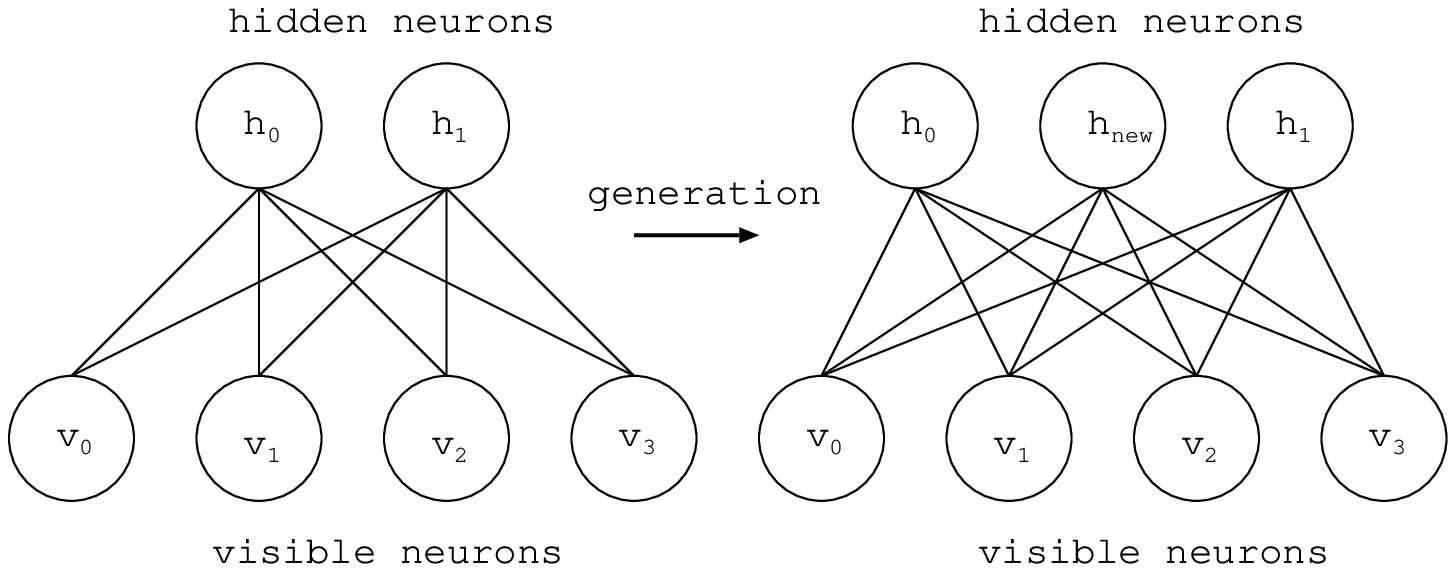}\label{fig:neuron_generation}}
\subfigure[Neuron annihilation]{\includegraphics[scale=0.5]{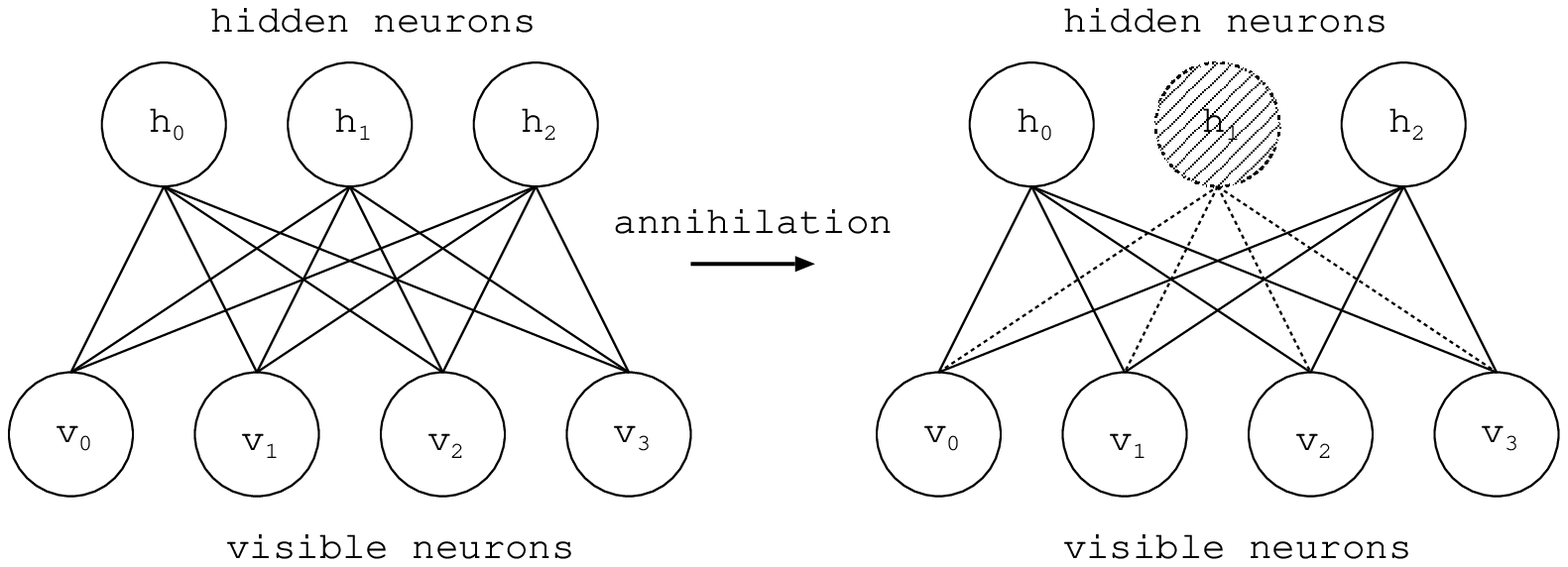}\label{fig:neuron_annihilation}}
\vspace{-3mm}
\caption{Adaptive RBM}
\label{fig:adaptive_rbm}
\end{center}
\end{figure}

The key idea of the neuron generation is Walking Distance (WD), which is inspired from the multi-layered neural network in the paper \cite{Ichimura95}. WD is the difference between the past variance and the current variance for learning parameters. The paper \cite{Ichimura95} described that if the network does not have enough neurons to classify them sufficiently, then the WD will tend to fluctuate large after the long training process. The situation shows that some hidden neurons may not represent an ambiguous pattern due to the lack of the number of hidden neurons. In order to represent ambiguous patterns into two neurons, a new neuron is inserted to inherit the attributes of the parent hidden neuron as shown in Fig.~\ref{fig:neuron_generation}.

In addition to the neuron generation, the neuron annihilation algorithm was applied to Adaptive RBM after neuron generation process. We had a situation that some unnecessary or redundant neurons were generated due to the neuron generation process. Therefore, such neurons can be removed according to the situation of output activation in the model after neuron generation process. Fig.~\ref{fig:neuron_annihilation} shows that the corresponding neuron is annihilated.

\begin{figure*}[]
\centering
\includegraphics[scale=0.8]{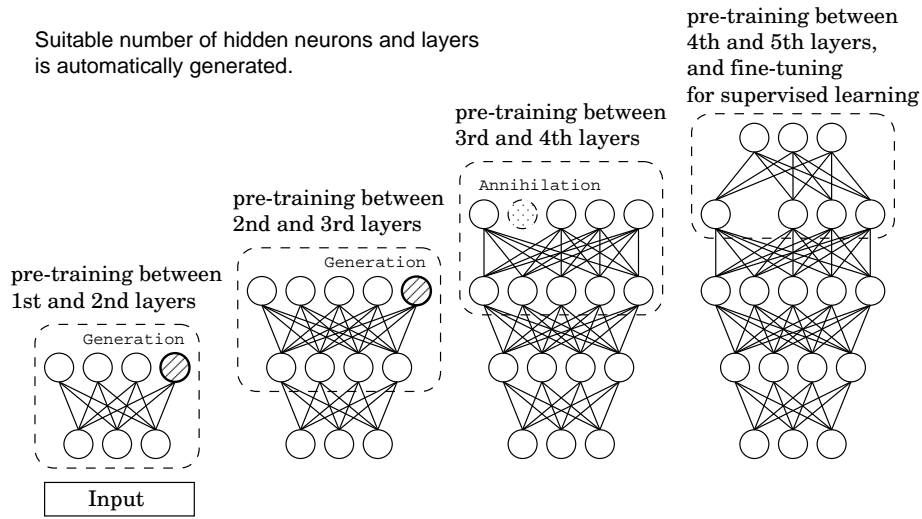}
\vspace{-3mm}
\caption{An Overview of Adaptive DBN}
\label{fig:adaptive_dbn}
\end{figure*}

\subsection{Layer Generation Algorithm of DBN}
\label{subsec:adaptive_dbn}
DBN is a hierarchical model of stacking the several pre-trained RBMs. For building process, output (hidden neurons activation) of $l$-th RBM can be seen as the next input of $l+1$-th RBM. Generally, DBN with multiple RBMs has higher data representation power than one RBM. Such hierarchical model can represent the specified features from an abstract concept to concrete representation at each layer in the direction to output layer. However, the optimal number of RBMs depends on the target data space.

We developed Adaptive DBN which can automatically adjust an optimal network structure by the self-organization by using the idea of WD. If the values of both WD and the energy function do not become small values, then a new RBM will be generated to keep the suitable network structure for the data set, since the RBM has lacked data representation capability to figure out an image of input patterns. Therefore, the condition for layer generation is defined by using the total WD and the energy function. Fig.~\ref{fig:adaptive_dbn} shows the overview of layer generation in Adaptive DBN.

\section{AffectNet and Facial Emotion models}
\label{sec:AffectNet}
AffectNet is a facial image database containing human emotions created by \cite{AffectNet}. As shown in Table \ref{tab:affectnet_category}, 11 categories are given and divided into training cases and test cases, respectively.

\begin{figure}[]
\centering
  \subfigure[Neutral]{\includegraphics[width=20mm]{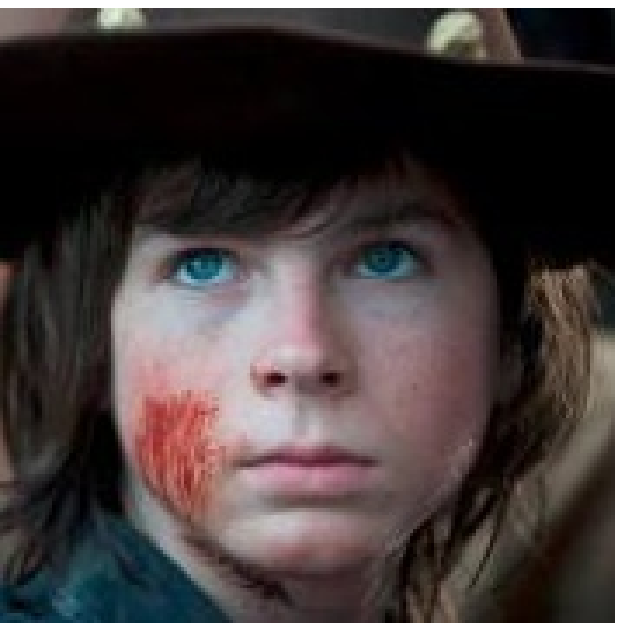}\label{fig:sample_0}}
  \subfigure[Happy]{\includegraphics[width=20mm]{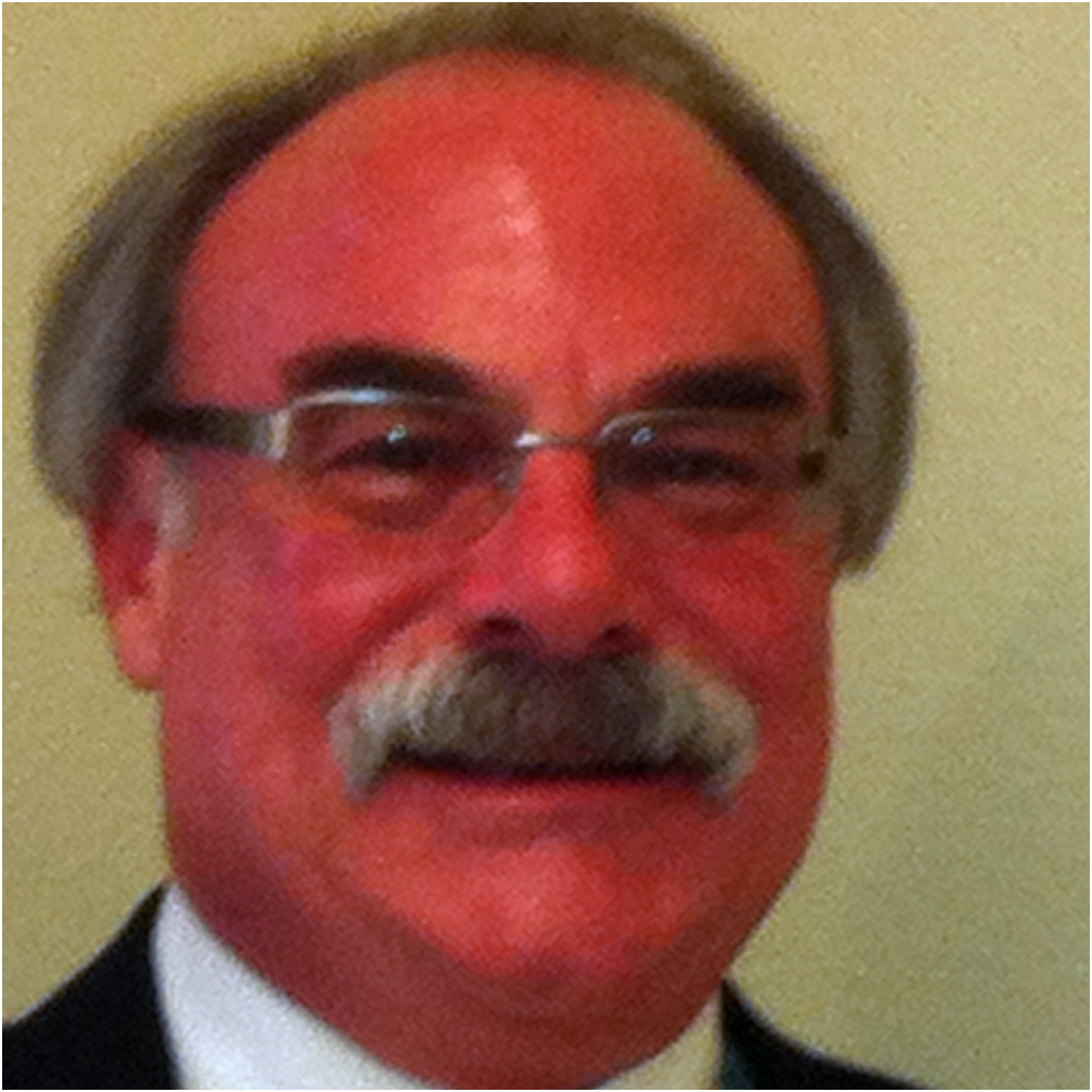}\label{fig:sample_1}}
  \subfigure[Sad]{\includegraphics[width=20mm]{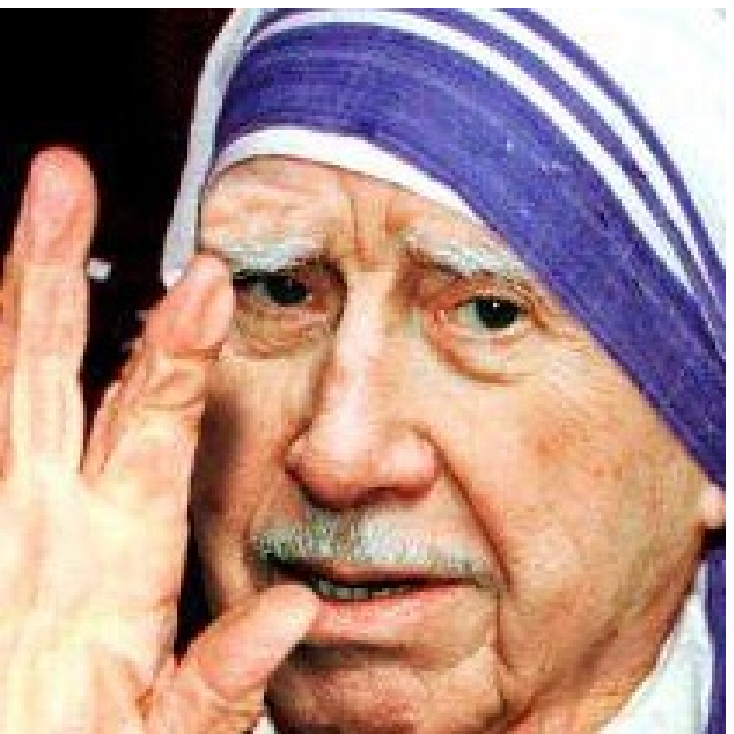}\label{fig:sample_2}}
  \subfigure[Surprise]{\includegraphics[width=20mm]{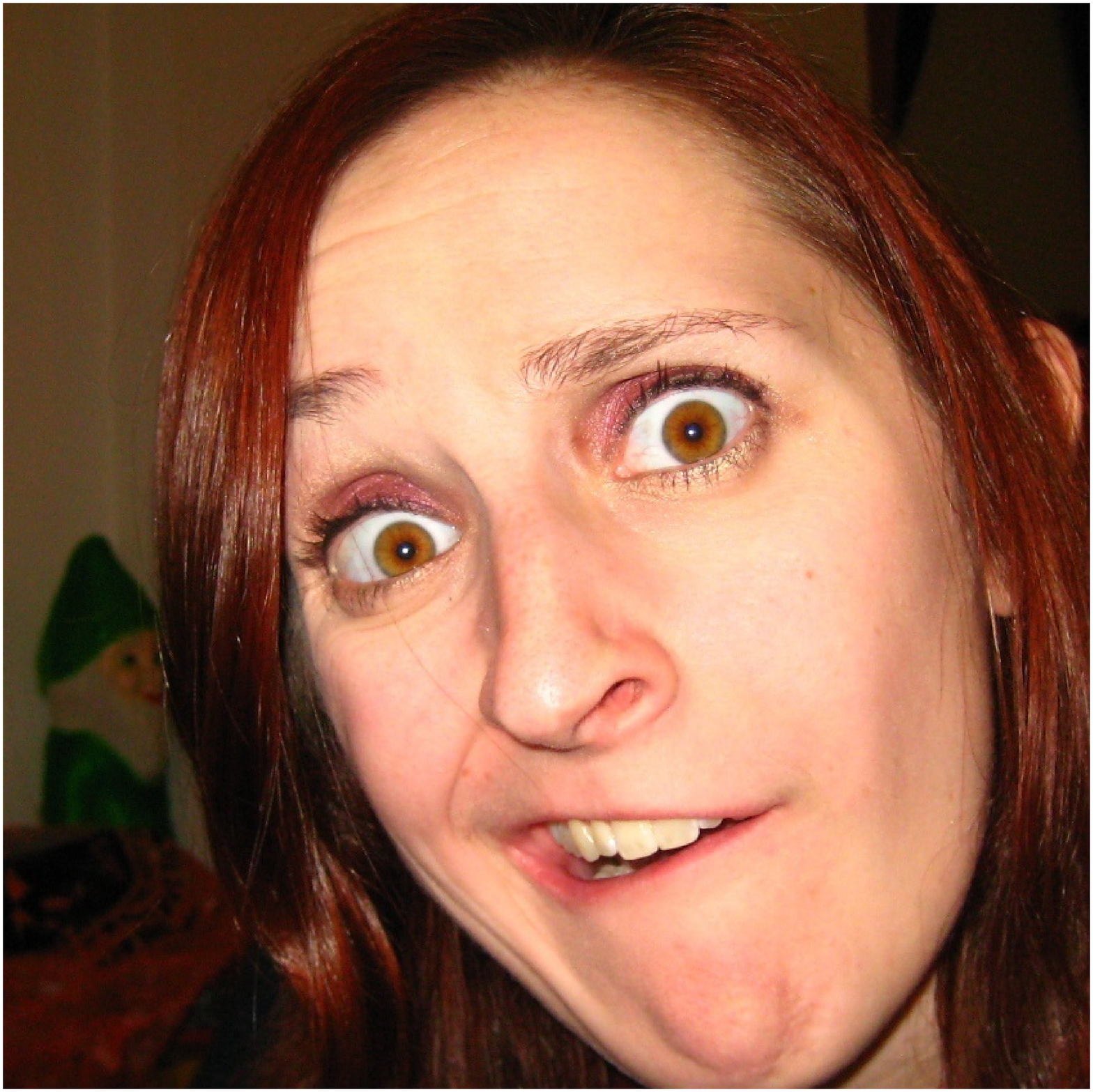}\label{fig:sample_3}}
  \subfigure[Fear]{\includegraphics[width=20mm]{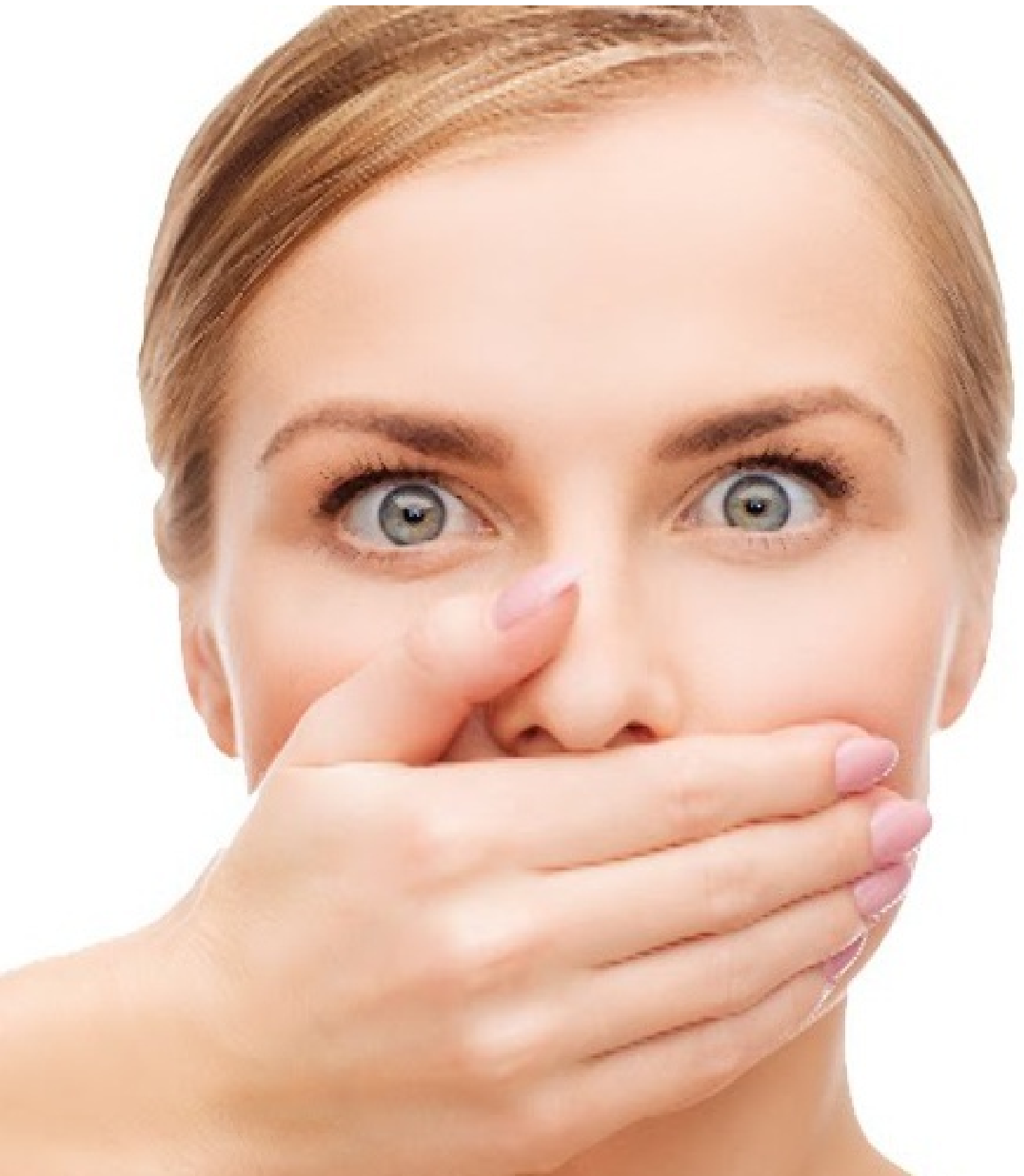}\label{fig:sample_4}}
  \subfigure[Disgust]{\includegraphics[width=20mm]{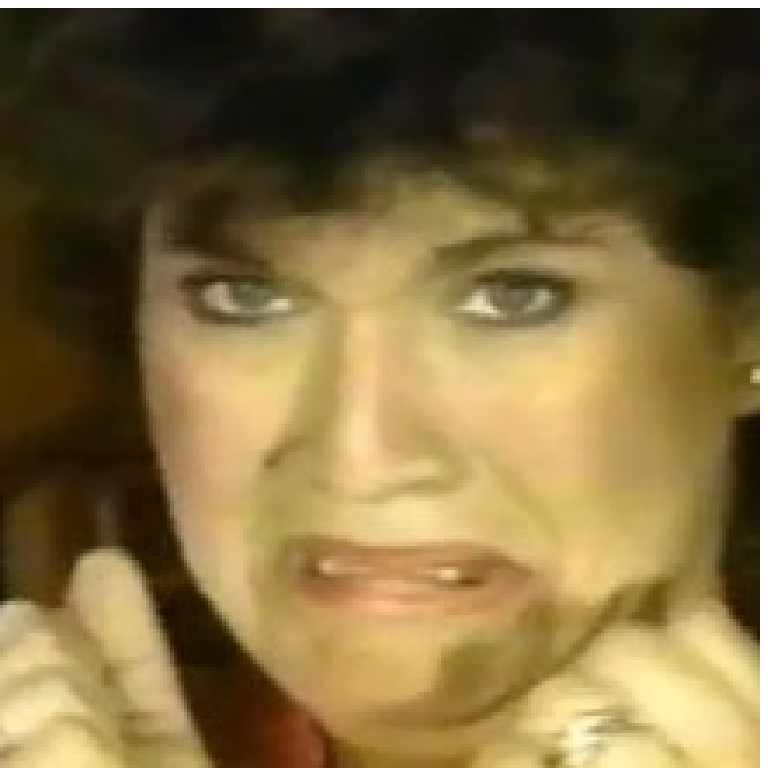}\label{fig:sample_5}}
  \subfigure[Anger]{\includegraphics[width=20mm]{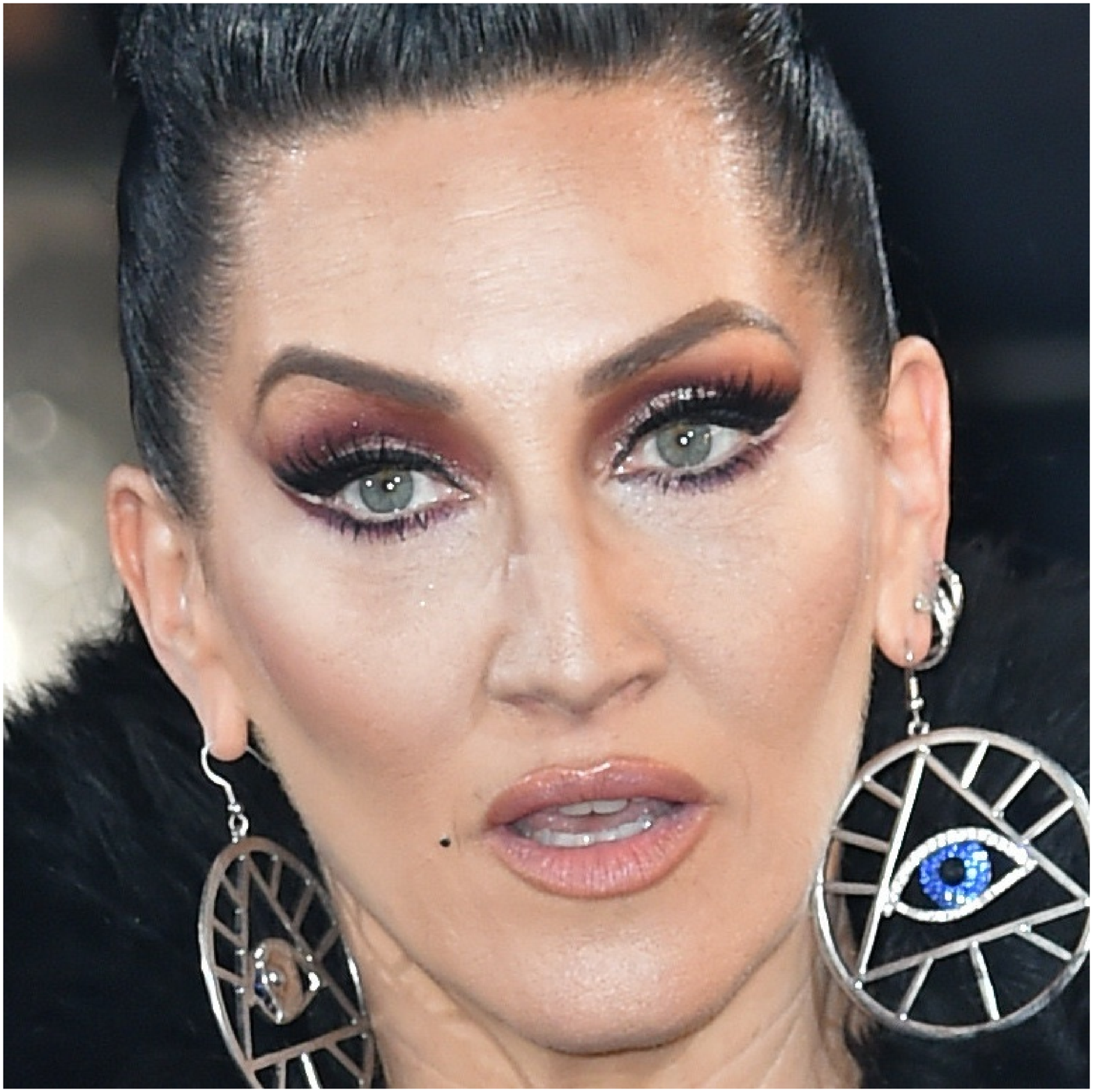}\label{fig:sample_6}}
  \subfigure[Contempt]{\includegraphics[width=20mm]{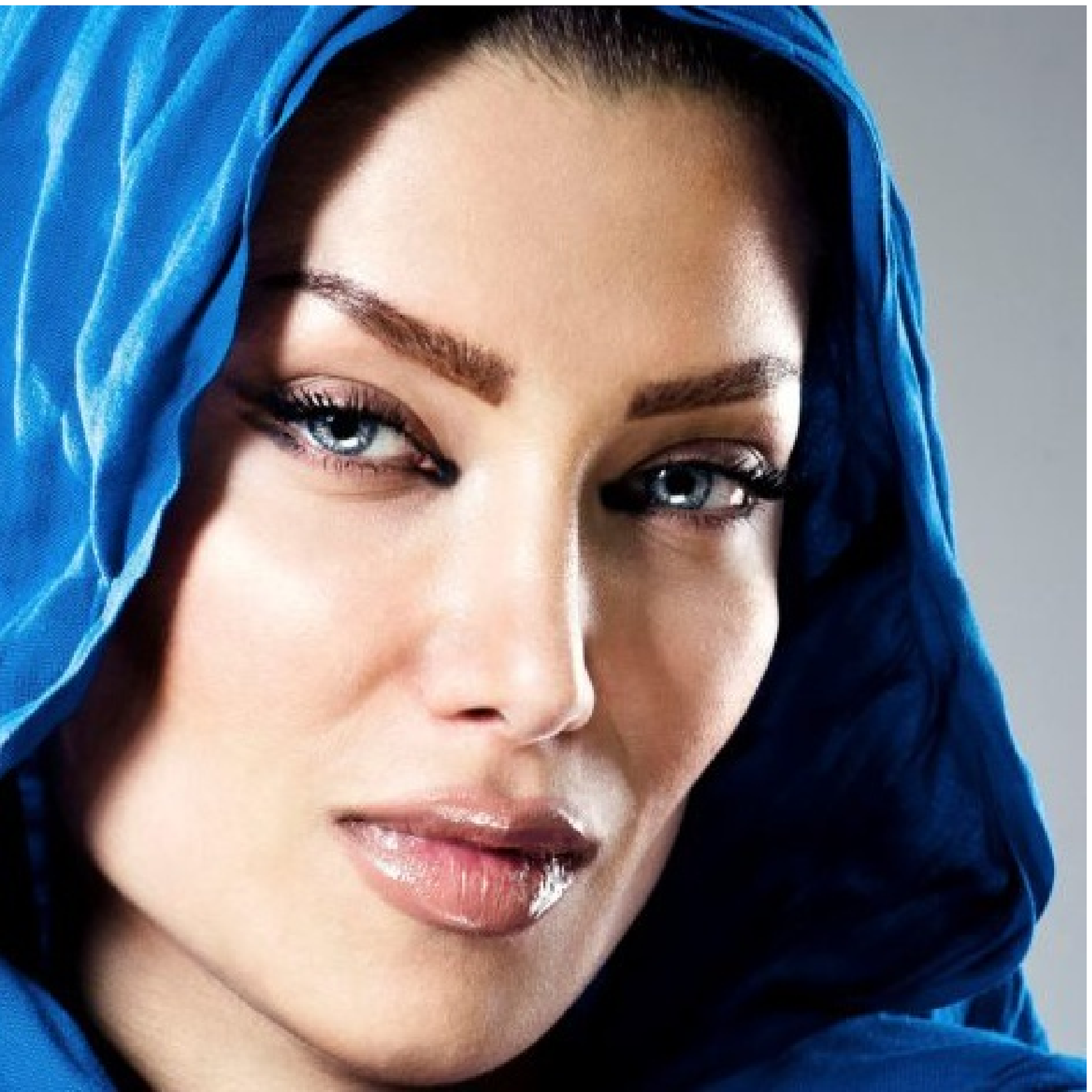}\label{fig:sample_7}}
  \subfigure[None]{\includegraphics[width=20mm]{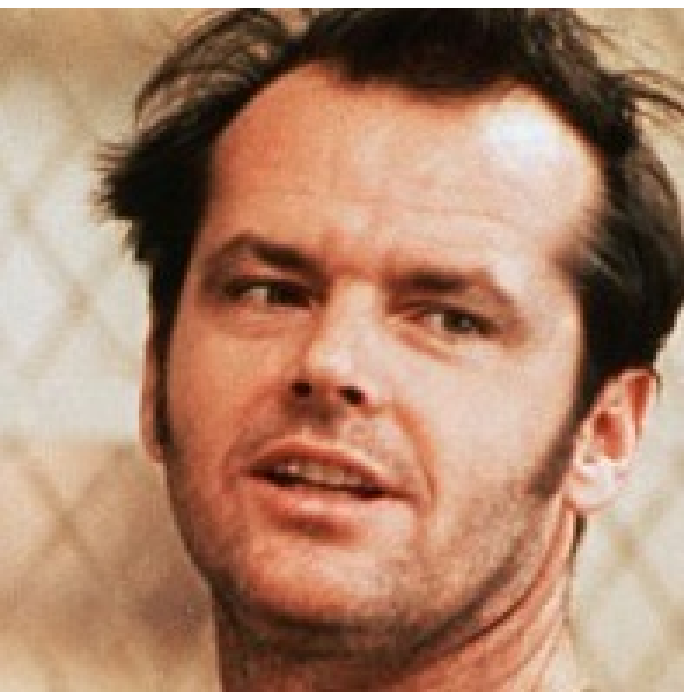}\label{fig:sample_8}}
  \subfigure[Uncertain]{\includegraphics[width=20mm]{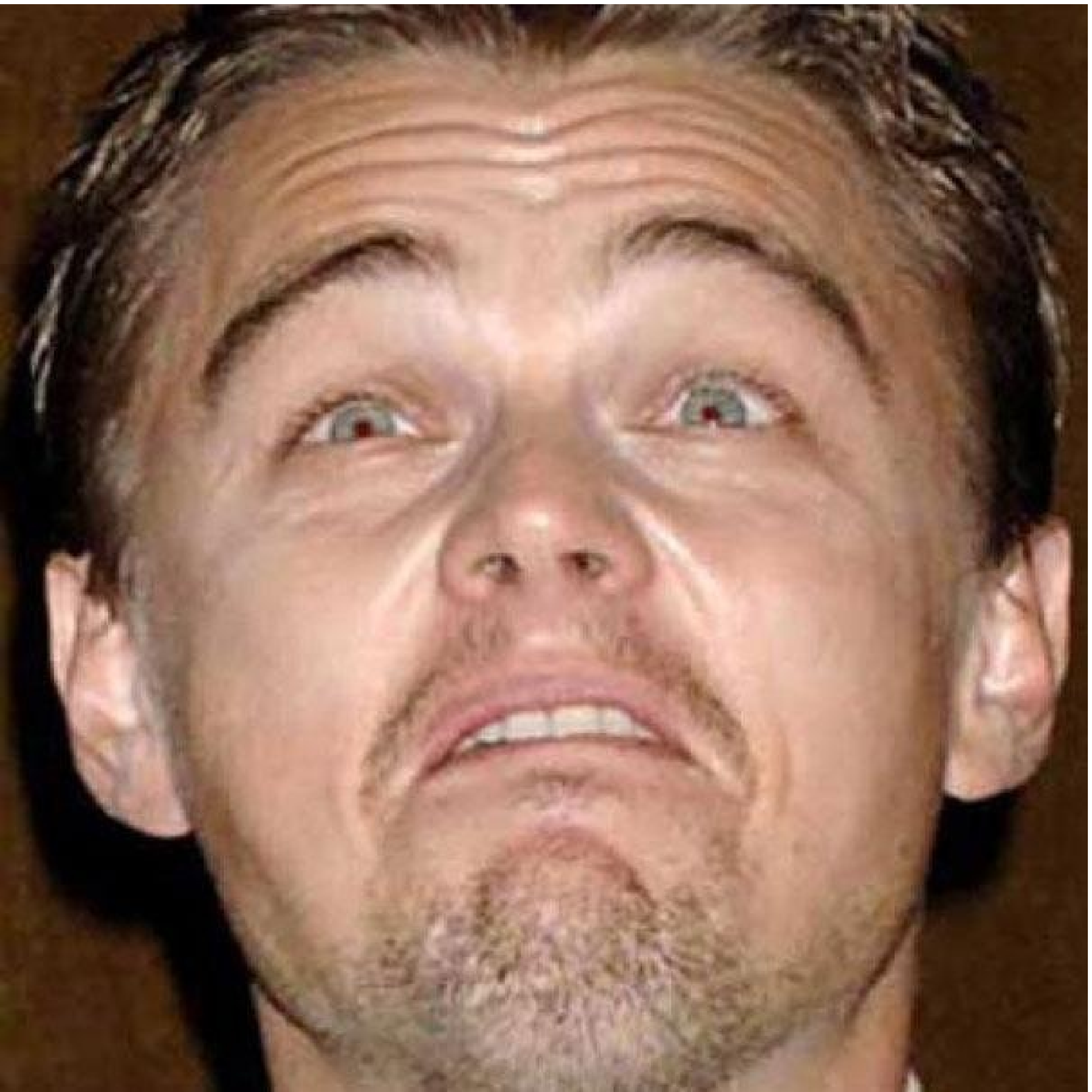}\label{fig:sample_9}}
  \subfigure[Non-Face]{\includegraphics[width=20mm]{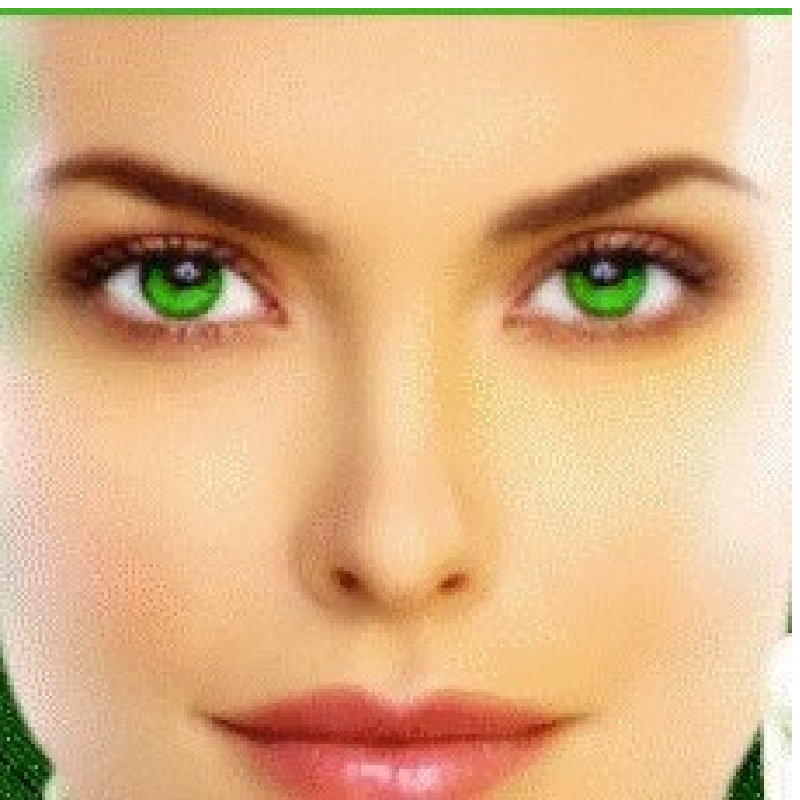}\label{fig:sample_10}}    
  \caption{Facial Images in AffectNet}
  \label{fig:affectnet_sample}
%\vspace{-5mm}
\end{figure}

Fig.~\ref{fig:affectnet_sample} shows the 11 categories of facial image. These categories are labeled according to human subjectivity based on valence and arousal to each facial image collected from the Internet. In \cite{AffectNet}, There is a description of the corresponding rate which two annotators make a same classification, that is, they respond a facial image as the same emotion category. The corresponding rate of two annotators' answer is not high, because human emotions contain many vague features and each annotator can be not always to perceive same emotion. Please refer \cite{AffectNet} for the difference of category.

\begin{figure}[]
  \centering
  \includegraphics[scale=0.4]{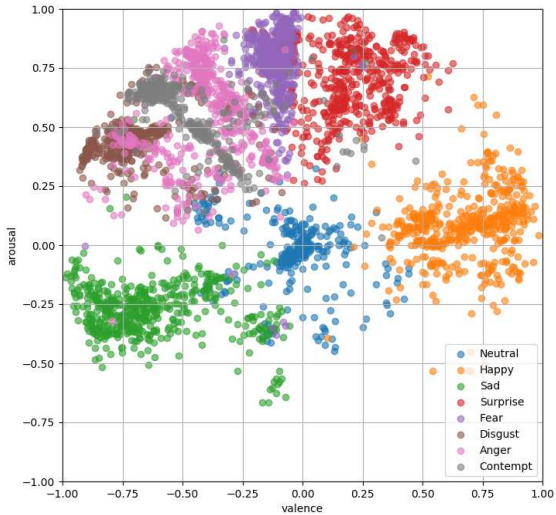}
\vspace{-3mm}
  \caption{Valence and Arousal}
  \label{fig:valence_arousal}
\end{figure}

The three categories `None', `Uncertain' and `Non-Face' are not categories related to facial emotions. This paper also excludes the three categories in the same way of \cite{AffectNet}. Moreover, the database contains the values ​​of valence and arousal, a rectangle indicating face in the image, and 68 face landmarks in an image. This paper uses valence and arousal in $[-1, 1]$ as shown in Fig.~\ref{fig:valence_arousal}. Fig.~\ref{fig:valence_arousal} shows the distribution of valence and arousal for eight emotion categories.

\begin{table}[]
\caption{The number of images in Each Category}
\vspace{-3mm}
\label{tab:affectnet_category}
\begin{center}
%\scalebox{0.8}[0.8]{
\begin{tabular}{l|r|r}
\hline 
\multicolumn{1}{c|}{Category} & Training & Test \\ \hline
Neutral   &  74,874  & 500 \\ 
Happy     & 134,415 &  500 \\ 
Sad       &  25,459 &  500 \\ 
Surprise  &  14,090 &  500 \\ 
Fear      &   6,378 &  500 \\ 
Disgust   &   3,803 &  500 \\ 
Anger     &  24,882 &  500 \\ 
Contempt  &   3,750 &  500 \\ 
None      &  33,088 &  500 \\ 
Uncertain &  11,645 &  500 \\ 
Non-Face  &  82,414 &  500 \\
\hline
\multicolumn{1}{c|}{Sum} & 414,798 & 5500\\ 
\hline
\end{tabular}
%} 
\end{center}
%\vspace{-5mm}
\end{table}

\section{Experimental Results}
\label{sec:Experiments}
\subsection{Classification Results}
\label{sec:Classification_Results}
The Adaptive DBN was trained for AffectNet. Table \ref{tab:classification_ratio} shows the classification results of the model. The most classified category was `Happy', while the most misclassified category was `Anger'. For test data, the classification accuracy for `Anger' was 78.4\%. The categories that classification accuracy was smaller than 90.0\% were `Neutral,' `Sad,' `Surprise,' `Anger,' and `Contempt.' Table \ref{tab:f1-score} shows the F1-score on AlexNet \cite{AffectNet} and Adaptive DBN. The macro average of Adaptive DBN was 0.874. The result of Adaptive DBN was better than that of AlexNet. F1-score means the harmonic average of precision and recall as follows.
\begin{eqnarray}
  Precision &=& \frac{TP}{TP+FP}, \\
  Recall &=& \frac{TP}{TP+FN}, \\
  F1 &=& 2\frac{P \times R}{P+R},
\end{eqnarray}
where TP is True Positive, FP is False Positive, and FN is False Negative. P is Precision and R is Recall.

\begin{table}[]
\caption{Classification Ratio by Adaptive DBN}
\vspace{-3mm}
\label{tab:classification_ratio}
\begin{center}
%\scalebox{0.9}[0.9]{
\vspace{-1mm}
\begin{tabular}{l|r|r}
\hline 
\multicolumn{1}{c|}{Category}  & Training set & Test set \\ 
\hline
Neutral   & 99.3\% &  87.8\% \\ 
Happy     & 99.9\% &  92.4\% \\ 
Sad       & 99.2\% &  84.2\% \\ 
Surprise  & 99.4\% &  85.8\% \\ 
Fear      & 99.5\% &  90.4\% \\ 
Disgust   & 99.3\% &  92.4\% \\ 
Anger     & 98.2\% &  78.4\% \\ 
Contempt  & 98.8\% &  87.6\% \\ \hline
Ave.      & 99.3\% &  87.4\% \\
\hline
\end{tabular}
%} 
\end{center}
%\vspace{-5mm}
\end{table}

\begin{table}[]
\caption{F1-score}
\vspace{-3mm}
\label{tab:f1-score}
\begin{center}
\vspace{-1mm}
%\scalebox{0.9}[0.9]{
\begin{tabular}{l|r|r}
\hline 
\multicolumn{1}{c|}{Category} & \multicolumn{1}{c|}{CNN\cite{AffectNet}} & Adaptive DBN \\
\hline
Neutral   & 0.63 & 0.85 \\
Happy     & 0.88 & 0.93 \\
Sad       & 0.63 & 0.87 \\
Surprise  & 0.61 & 0.88 \\
Fear      & 0.52 & 0.90 \\
Disgust   & 0.52 & 0.83 \\
Anger     & 0.65 & 0.85 \\
Contempt  & 0.08 & 0.85 \\ \hline
Ave.      & - & 0.874 \\
\hline
\end{tabular}
%} 
\end{center}
%\vspace{-5mm}
\end{table}

\begin{table*}[]
\caption{Confusion Matrix for Classification Results}
\vspace{-3mm}
\label{tab:classification_ratio_cf}
\begin{center}
%\scalebox{0.8}[0.8]{
\begin{tabular}{l|l|r|r|r|r|r|r|r|r}
\hline
\multicolumn{2}{c|}{} & \multicolumn{8}{c}{Predicted Category} \\ \cline{3-10}
\multicolumn{2}{c|}{} & Neutral & Happy & Sad & Surprise & Fear & Disgust & Anger & Contempt \\ \hline 
\multirow{8}{*}{\rotatebox[origin=c]{90}{Real Category}}& Neutral & 439 & 2 & 7 & 5 & 8 & 16 & 4 & 19 \\
& Happy & 7 & 462 & 2 & 0 & 4 & 12 & 1 & 12 \\
& Sad & 12 & 3 & 421 & 13 & 11 & 20 & 5 & 15 \\
& Surprise & 15 & 4 & 10 & 429 & 11 & 22 & 0 & 9 \\
& Fear & 10 & 2 & 10 & 10 & 452 & 8 & 3 & 5 \\
& Disgust & 8 & 2 & 3 & 5 & 8 & 462 & 5 & 7 \\
& Anger & 14 & 4 & 8 & 10 & 9 & 47 & 392 & 16 \\
& Contempt & 17 & 8 & 6 & 3 & 2 & 21 & 5 & 438 \\
\hline 
\end{tabular}
%} 
\end{center}
%\vspace{-5mm}
\end{table*}

\begin{figure}[h]
  \centering
  \subfigure[Disgust $\rightarrow$ Anger]{\includegraphics[width=25mm]{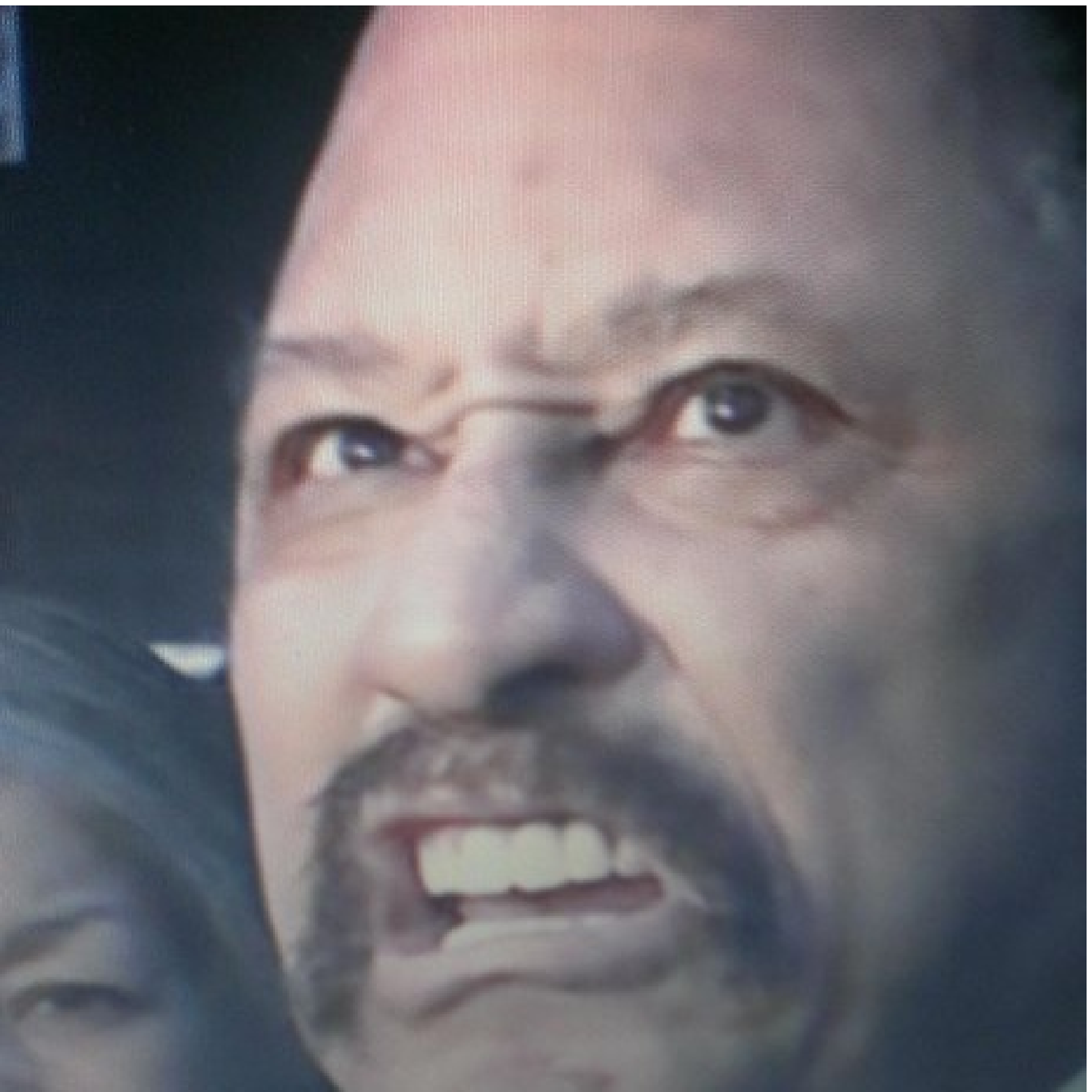}\label{fig:WA1}}
  \subfigure[Disgust $\rightarrow$ Anger]{\includegraphics[width=25mm]{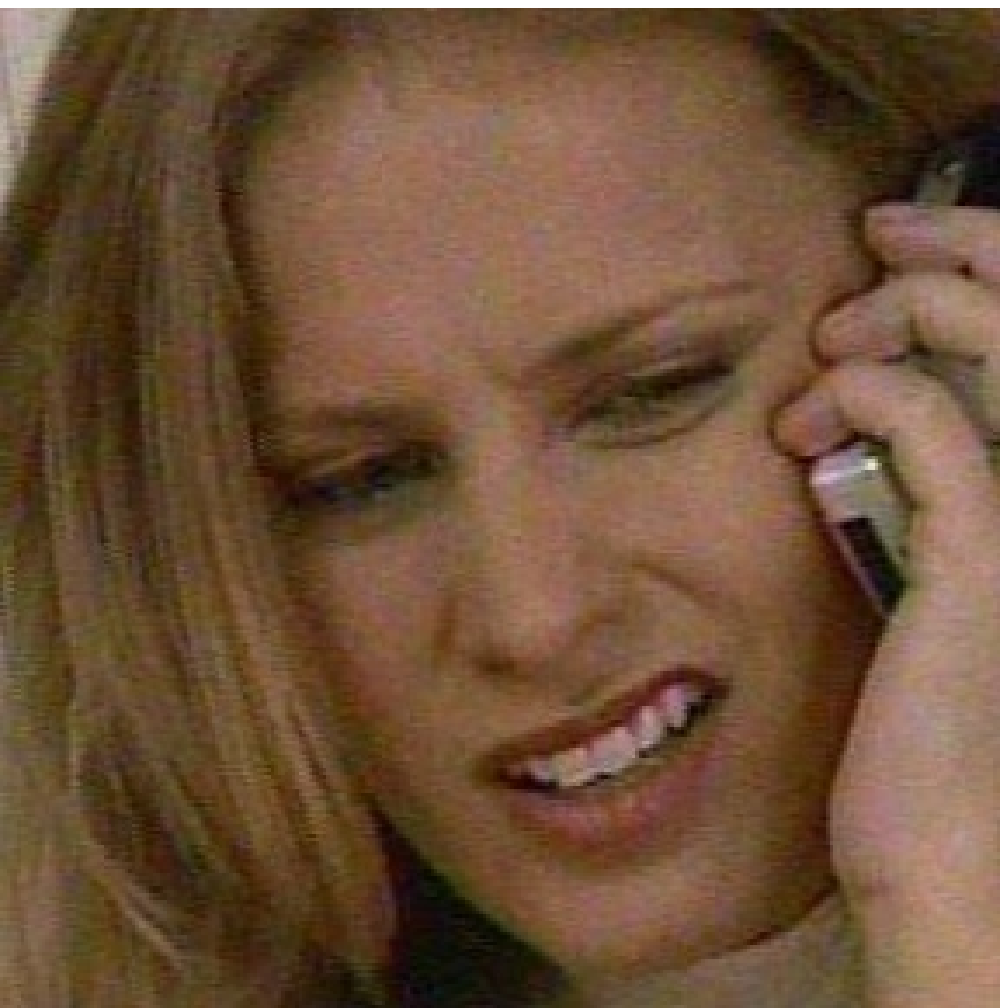}\label{fig:WA2}}
  \subfigure[Disgust $\rightarrow$ Anger]{\includegraphics[width=25mm]{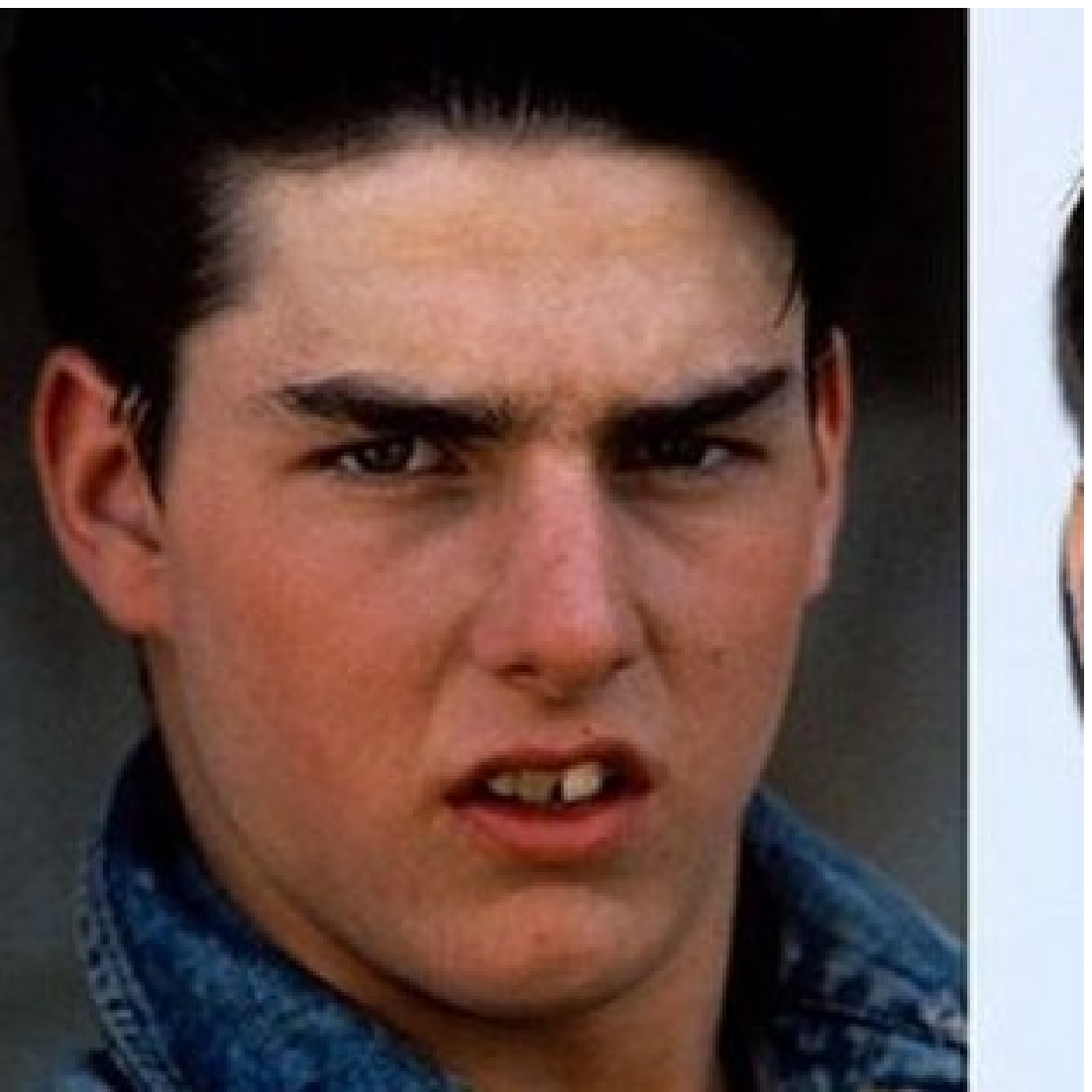}\label{fig:WA3}}
  \subfigure[Anger $\rightarrow$ Disgust]{\includegraphics[width=25mm]{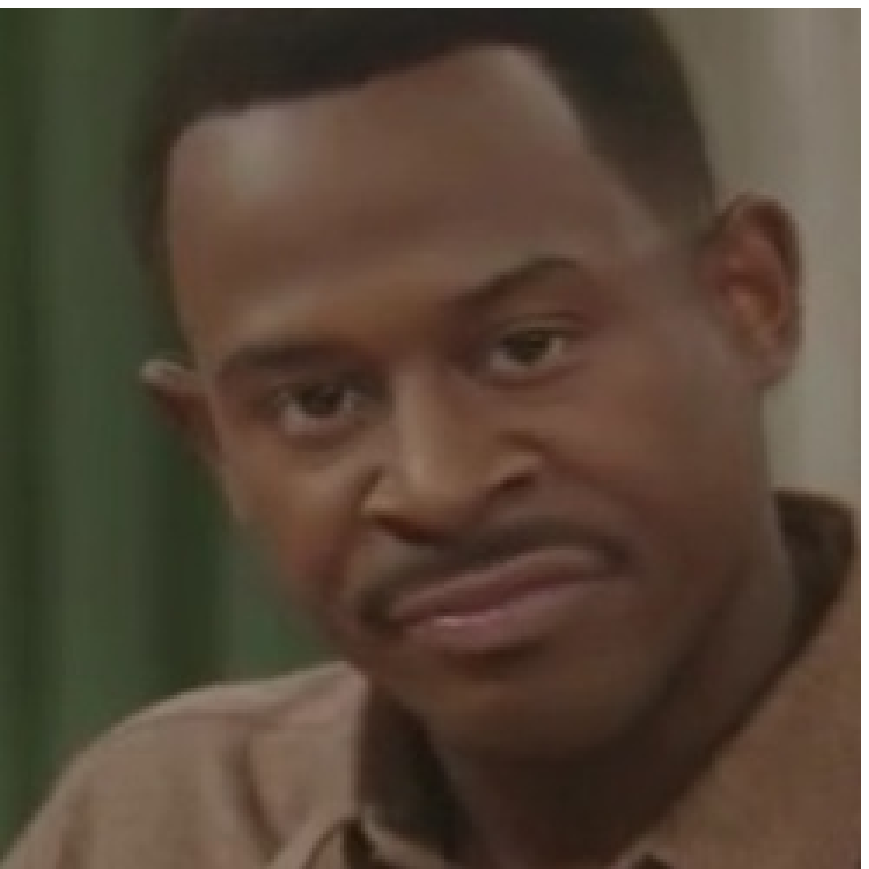}\label{fig:WA4}}
  \subfigure[Anger $\rightarrow$ Disgust]{\includegraphics[width=25mm]{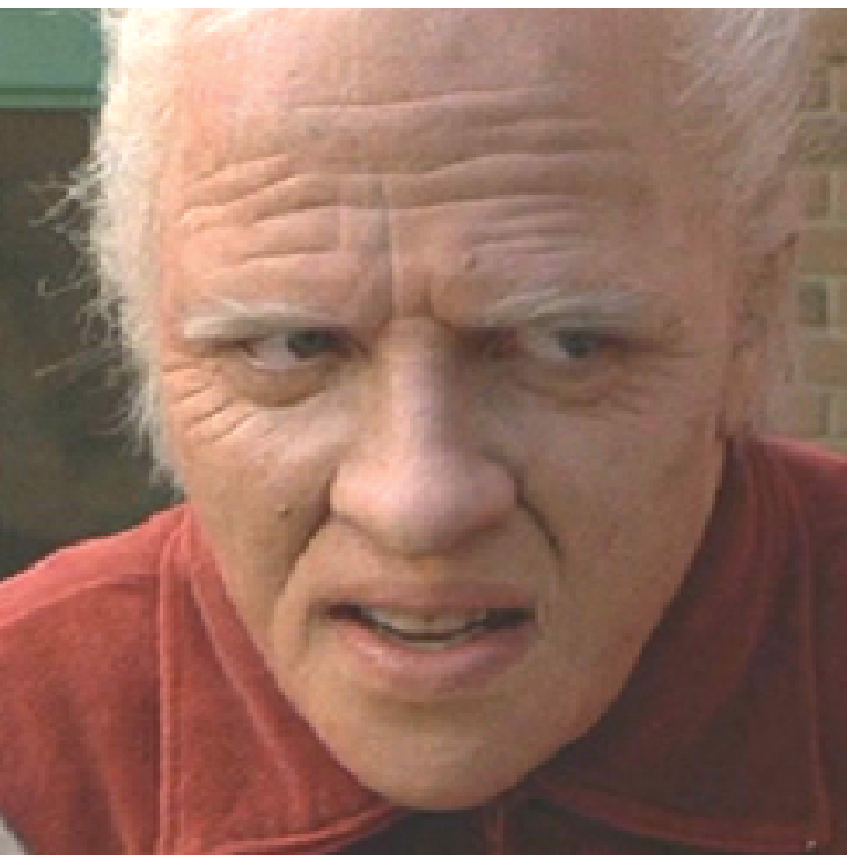}\label{fig:WA5}}
  \subfigure[Anger $\rightarrow$ Disgust]{\includegraphics[width=25mm]{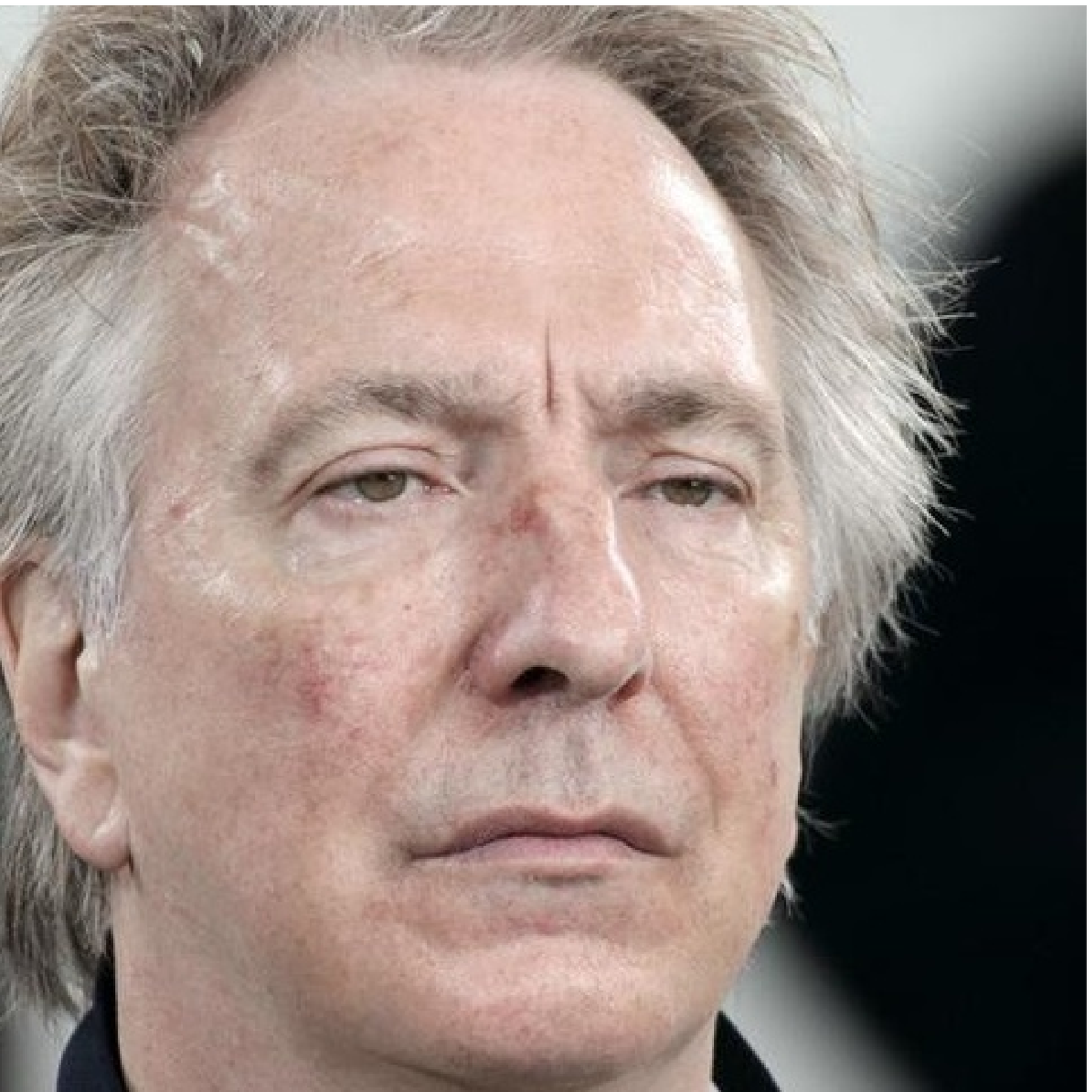}\label{fig:WA6}}
  \caption{Wrong answer cases}
  \label{fig:sample_wrong}
%\vspace{-5mm}
\end{figure}

We investigated the predicted category for the misclassified cases. Table \ref{tab:classification_ratio_cf} shows the confusion matrix for the classification results in Table \ref{tab:classification_ratio}. The confusion matrix means the correlation table of the predicted category and the true category for facial images. As shown in Table \ref{tab:classification_ratio_cf}, `Anger' is misclassified as `Disgust'. Fig.~\ref{fig:sample_wrong} shows the wrong answer cases. As shown in Fig.~\ref{fig:WA1} to \ref{fig:WA3}, there are wrong cases that the model answers `Anger', but the correct is `Disgust' category. Fig.~\ref{fig:WA4} to \ref{fig:WA6} are the contrary cases.

\subsection{Child model and Its KL Divergence}
\label{sec:KL_Divergence}
The difference between `Anger' and `Disgust' was not able to be distinguished from the experimental results in Table \ref{tab:classification_ratio}. The paper \cite{AffectNet} discusses that two or more annotators answer different subjective judgment, this is, the answer depends on the person. Fig.~\ref{fig:valence_arousal} shows that there is overlap with respect to `Anger' and `Disgust' in the 2D space of valence and arousal.

In order to distinguish such cases leading wrong answer, to use two or more child models is effective to represent them. As opposed to general ensemble learning, we have already built the Adaptive DBN model with high classification capability in Section \ref{sec:Classification_Results}. Therefore, some additional child models are added to the original parent model for re-learning. Some appropriate child models are generated according to difference between two models by using KL divergence.

In this paper, we focused only two categories, `Anger' and `Disgust', which is emarkable difference from reported results in \cite{AffectNet}, because the other emotions were correctly classified by the parent model as mentioned in Section \ref{sec:Classification_Results}. The data on `Anger' and `Disgust' were trained by using two child models, and KL divergence showed the difference in distribution between the models. The valence and arousal for facial images with large KL were investigated. The model in Section \ref{sec:Classification_Results} was defined as the parent model $P$. Table \ref{tab:relearning_dataset} shows the data set for the evaluation of child models. Let $Q1$ and $Q2$ be a child model to train Set 1 and Set 2, respectively. $Q1$ and $Q2$ for $P$ was compared by using KL divergence which is defined as the following equation.
\begin{equation}
\label{eq:kl_f}
D_{KL}(P, Q) = \sum_{i} P(x_{i}) \log \frac{P(x_{i})}{Q(x_{i})},
\end{equation}
where $P$ and $Q$ are distribution estimated by softmax layer for parent model and child model, respectively. $x_{i}$ is an input signal. $D_{KL}(P, Q)$ means the KL divergence for $x_{i}$.

\begin{figure}[]
  \centering
  \includegraphics[scale=0.5]{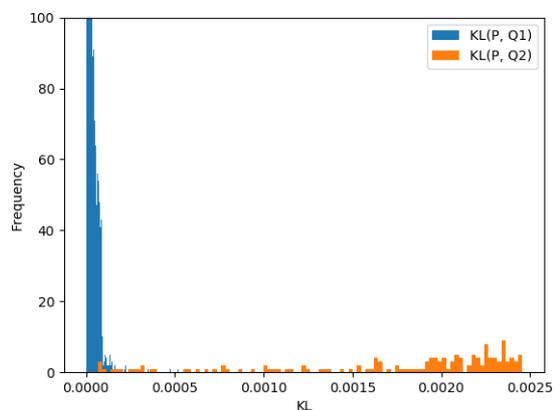}
\vspace{-5mm}
  \caption{Difference between KL(P, Q1) and KL(P, Q2)}
  \label{fig:kl-p-q1-q2}
\end{figure}

Table \ref{tab:kl} shows the KL divergence with Eq.~(\ref{eq:kl_f}): $KL(P, Q1)$ and $KL(P, Q2)$ for given input. As a result, $D_{KL}(P, Q2)$ was larger than $D_{KL}(P, Q1)$, this is, $Q2$ model represented the wrong answer. Fig.~\ref{fig:kl-p-q1-q2} also shows similar result, but it is histogram of KL for each sample. Vertical axis and horizontal axis in Fig.~\ref{fig:kl-p-q1-q2} are the value of KL and frequency, respectively. As shown in Fig.~\ref{fig:kl-p-q1-q2}, the value of $D_{KL}(P, Q1)$ is almost $0.0000$, while the value of $D_{KL}(P, Q2)$ has wide range with $[0.0000, 0.0025]$. The results showed the misclassified samples were represented according to the value of KL divergence.

\begin{figure}[]
  \centering
  \subfigure[$\theta_{KL} = 0.0010$]{\includegraphics[scale=0.5]{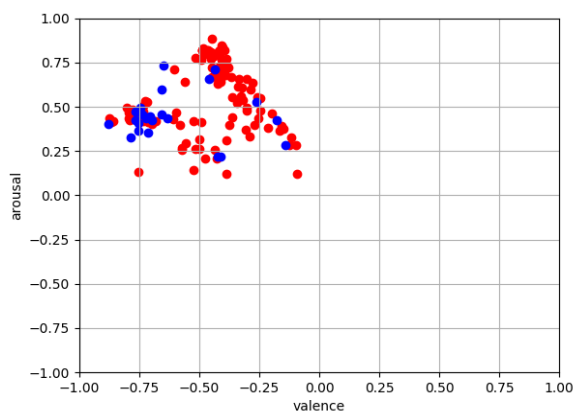}\label{fig:kl_0.0010}\vspace{-3mm}}
  \subfigure[$\theta_{KL} = 0.0015$]{\includegraphics[scale=0.5]{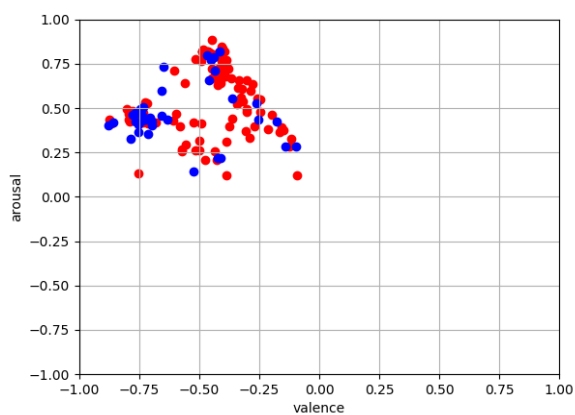}\label{fig:kl_0.0015}\vspace{-3mm}}
  \subfigure[$\theta_{KL} = 0.0020$]{\includegraphics[scale=0.5]{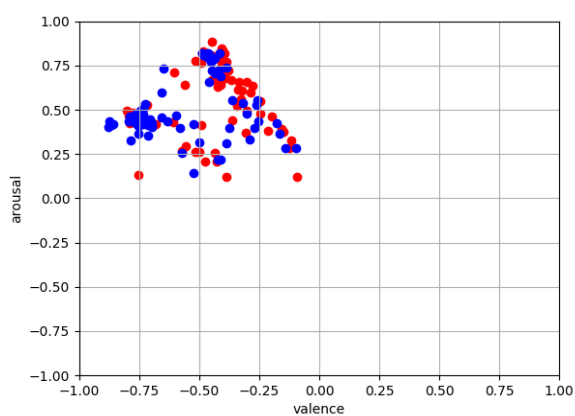}\label{fig:kl_0.0020}\vspace{-3mm}}
  \caption{Valence and Arousal in KL(P, Q2)}
  \label{fig:kl-valence_arousal}
%\vspace{-5mm}
\end{figure}

According to the consideration of the KL distribution as shown in Fig.~\ref{fig:kl-p-q1-q2} and the classification results in Table \ref{tab:kl}, we make an assumption that the samples are divided by $\theta_{KL} $ as $\{0.0020, 0.0015, 0.0010\}$. The samples with KL thresholds in Set 2 are plotted in valence and arousal 2D space. As shown in Fig.~\ref{fig:kl-valence_arousal}, red / blue circle is the sample with larger / smaller than KL threshold. Table \ref{tab:result_kl} shows the classification ratio of the re-learning child model with each KL threshold. As a result, in case of $\theta_{KL} = 0.0015$, the child model was better performance. From such observations, the deep learning model classified the facial data correctly, even for the wrong cases including two distributed categories by mixture. Good child model requires the appropriate segmentation of data distribution by using KL divergence and then we will give its condition of trade-off between KL divergence and the classification capability of child models in near future.

\begin{table}[]
\caption{Data set for re-learning child model}
\vspace{-3mm}
\label{tab:relearning_dataset}
\begin{center}
%\scalebox{0.8}[0.8]{
\begin{tabular}{c|l|r}
\hline 
dataset & \multicolumn{1}{c|}{description} & cases  \\ 
\hline
Set 0 & All data in Anger and Disgust               &   1,000 \\ 
Set 1 & Correct cases for Anger and Disgust. $Q1$ trains Set 1.  &  854 \\ 
Set 2 & Wrong case for Anger and Disgust. $Q2$ trains Set 2. &  146 \\ 
\hline
\end{tabular}
%} 
\end{center}
%\vspace{-5mm}
\end{table}

\begin{table}[]
\caption{KL Divergence}
\vspace{-3mm}
\label{tab:kl}
\begin{center}
\begin{tabular}{l|r}
\hline 
\multicolumn{1}{c|}{Model} & KL  \\ \hline
$D_{KL}(P, Q1)$   & 0.188 \\
$D_{KL}(P, Q2)$   & 0.660 \\ 
\hline
\end{tabular}
\end{center}
\end{table}

\begin{table}[]
  \caption{Classification ratio of re-learning child model}
\vspace{-3mm}
\label{tab:result_kl}
\begin{center}
\begin{tabular}{c|r}
\hline 
$\theta_{KL}$ & \multicolumn{1}{c}{Classification ratio} \\ \hline
0.20 & 95.8\% \\
0.15 & 97.2\% \\
0.10 & 95.2\% \\
\hline
\end{tabular}
\end{center}
\end{table}

\section{Conclusion}
\label{sec:Conclusion}
Adaptive structural learning method of DBN has high classification capability for large image benchmark databases. In this paper, the facial image database: AffectNet was trained by Adaptive DBN. However, the deep learning model of test dataset was not able to realize perfect classification for the specified emotional categories. This paper investigated the KL divergence of Adaptive DBN models where the trained model is a parent model and the new trained model only for some cases at misclassified categories is a child model. From experimental results in this paper, we will discover a suitable threshold on the deep learning model with two or more child models by using KL divergence in future.

\section*{Acknowledgment}
This work was supported by JSPS KAKENHI Grant Number 19K12142, 19K24365, and obtained from the commissioned research by National Institute of Information and Communications Technology (NICT, 21405), JAPAN.

\end{document}